\documentclass[pdflatex,sn-mathphys-num,iicol]{sn-jnl}

\usepackage{booktabs}
\usepackage{amsmath}
\usepackage{amssymb}
\usepackage{graphicx}
\usepackage{xcolor}
\usepackage{tikz}
\usetikzlibrary{shapes.geometric,shapes.misc,arrows.meta,positioning,calc,shadows.blur}

\definecolor{BlueGray}{RGB}{52,73,94}       
\definecolor{LightBlue}{RGB}{214,229,245}   
\definecolor{GreenDark}{RGB}{39,119,72}     
\definecolor{GreenLight}{RGB}{213,241,222}  
\definecolor{RedDark}{RGB}{169,50,38}       
\definecolor{RedLight}{RGB}{250,219,216}    
\definecolor{OrangeDark}{RGB}{175,96,26}    
\definecolor{OrangeLight}{RGB}{253,235,208} 
\definecolor{FeedbackRed}{RGB}{192,57,43}   

\tikzset{
  proc/.style  = {rectangle, rounded corners=3pt,
                  draw=BlueGray, line width=1.1pt,
                  fill=LightBlue,
                  minimum width=3.4cm, minimum height=1.0cm,
                  align=center, font=\small\bfseries\color{BlueGray}},
  engbox/.style = {rectangle, rounded corners=3pt,
                   draw=BlueGray, line width=1.1pt,
                   fill=LightBlue,
                   minimum width=9.0cm, minimum height=1.0cm,
                   align=center, font=\small\bfseries\color{BlueGray}},
  vbox/.style  = {rectangle, rounded corners=2pt,
                  draw=OrangeDark, line width=0.7pt,
                  fill=OrangeLight,
                  minimum width=1.55cm, minimum height=0.7cm,
                  align=center, font=\scriptsize\color{OrangeDark}},
  certbox/.style = {rounded rectangle, rounded rectangle arc length=180,
                    draw=GreenDark, line width=1.1pt,
                    fill=GreenLight,
                    minimum width=2.6cm, minimum height=0.85cm,
                    align=center, font=\small\bfseries\color{GreenDark}},
  impbox/.style  = {rounded rectangle, rounded rectangle arc length=180,
                    draw=RedDark, line width=1.1pt,
                    fill=RedLight,
                    minimum width=2.8cm, minimum height=0.85cm,
                    align=center, font=\small\bfseries\color{RedDark}},
  cnode/.style = {circle, draw=black, thick, minimum size=0.95cm,
                  align=center, font=\small},
  onode/.style = {circle, draw=black, thick, double, minimum size=0.95cm,
                  align=center, font=\small},
  arr/.style     = {-{Latex[length=5pt,width=4pt]}, thick, draw=BlueGray},
  arrgreen/.style= {-{Latex[length=5pt,width=4pt]}, thick, draw=GreenDark},
  arred/.style   = {-{Latex[length=5pt,width=4pt]}, thick, draw=RedDark},
  arrfb/.style   = {-{Latex[length=5pt,width=4pt]}, line width=1.2pt,
                    draw=FeedbackRed},
  arrthin/.style = {-{Latex[length=4pt,width=3pt]}, thin},
  arrdash/.style = {-{Latex[length=5pt,width=4pt]}, thick, dashed},
}

\makeatletter
\long\def\@tablecaption#1#2{%
  \setbox\tabcapbox\vbox{\tablecaptionfont
  {\bfseries #1}{\hskip2mm}#2\vphantom{y}\par}%
  \box\tabcapbox%
}
\makeatother

\hypersetup{
  pdftitle={Structural Certification for Reliable Physical Design with Language Models},
  pdfauthor={Nakul Vyas and Iliya D. Stoev},
  pdfsubject={Scientific manuscript on structural certification for physical design with language models},
  pdfkeywords={machine reasoning, physical law, formal verification, self-correction, deterministic systems, propose-certify loop}
}

\begin{document}

\title{Structural Certification for Reliable Physical Design with Language Models}

\author*[1]{\fnm{Nakul} \sur{Vyas}}\email{nvyas@heysuvi.com}
\author*[1,2]{\fnm{Iliya} \sur{D. Stoev}}\email{istoev@heysuvi.com; iliya.stoev@kit.edu}

\affil[1]{\orgname{Heysuvi Labs, LLC}, \orgdiv{US Operations \& Research}, \orgaddress{\street{7345 W Sand Lake Rd, Ste 210}, \city{Orlando}, \postcode{32819}, \state{FL}, \country{United States}}}
\affil[2]{\orgname{Institute of Biological and Chemical Systems - Functional Molecular Systems, Karlsruhe Institute of Technology},
\orgaddress{\street{Hermann-von-Helmholtz-Platz 1}, \city{Eggenstein-Leopoldshafen}, \postcode{76344}, \state{Baden-Württemberg}, \country{Germany}}}

\abstract{An unreliable language model can be made to produce reliable physical designs if the authority to assert is moved out of the model: the model proposes, and a deterministic engine alone certifies, returning certified, impossible, or unknown. We introduce Physics-Anchored Certification (PHACT), a propose--certify loop spanning five scientific domains, and identify what makes such a certificate trustworthy. A checker that accepts a model-supplied value can be forged; deriving the certified quantity from fixed inputs instead makes forgery impossible by construction. Across eighty adversarial trials spanning two models, two decoding temperatures, and a deliberately faulted engine, this contract produced zero false certifications.}


\maketitle

\section*{Main}

A language model can produce a physically invalid answer with the same fluency and confidence as a valid one. In scientific and engineering settings, where a wrong result carries real experimental or computational cost, the problem is not only generation but certification: validity must be decided by an external physical check the model cannot author.

The usual response is to improve the predictor: more parameters, more data, better alignment. This helps but cannot resolve the problem, because the problem is not a deficit of knowledge. A model that has memorized every textbook will still, pushed off its training distribution, produce a confident wrong answer. The deficit is structural: the model is unable to decline. We take a different route. We do not make the model more reliable; we remove its authority to assert. We place it inside a loop with a deterministic physics engine (the model proposes, the engine disposes) so that a proposed design becomes an answer only once the engine has computed, from first principles, that it satisfies the governing law. On rejection the model is told which law was violated and by how much, and it revises. If every genuine attempt fails for the same physical reason, the model does what a bare model is never permitted to do: it reports that the goal is impossible.

A design is reliable to the extent the certifier can check it. We instantiate PHACT in five domains spanning aerodynamics, DNA thermodynamics, astrophysics, analog electronics, and colloidal drug formulation, to show the architecture is not tied to a single physical system. The proposer may be a hosted model, but the certifier is ordinary deterministic code. The certification step therefore runs on commodity hardware and can screen a candidate design against the governing physics before experimental effort or an expensive HPC allocation is committed. We release the full system to that end.

That openness matters only if the certificate itself can be trusted. The central contribution of this paper is identifying the structural condition a propose--certify loop must satisfy to make that possible. It is not enough to put a checker after the model. If the checker asks the model to \emph{supply} the quantity being certified (``here are the masses; confirm the chirp mass is $50\,M_\odot$''), then on a goal that should be refused, the model can quietly change a quantity the prompt declared fixed, hand the checker a self-consistent but off-goal design, and receive a certificate for correct physics that does not answer the question asked. We show this forgery is common, not hypothetical. The fix is structural: the checker must \emph{derive} the certified quantity from bound inputs rather than accept it from the model. We formalize this with a structural contract for each domain: a directed dependency graph in which quantities fixed by the goal enter as bound input nodes, any free design variables occupy separate writable nodes, and the certified quantity is a derived leaf the model cannot write. On a fixed-system verification goal, the bound nodes are read from the goal and are not writable by the model, so the model cannot substitute a new mass pair (for this astrophysics example, a different $m_1, m_2$) and ask the checker to certify a different system. Under this contract the forgery is impossible by construction, and we confirm it holds across models, decoding temperatures, and even when the surrounding engine is deliberately faulted.

\paragraph{Contributions}
\textbf{(i)} We formulate physically grounded design as a propose--certify loop in which a deterministic engine, not the model, holds the authority to assert, with a three-valued output (certified, impossible, unknown) in which declining is a valid verdict, not a failure mode.
\textbf{(ii)} We identify the condition under which a certificate can be forged, namely a validator that lets the model supply the quantity under test, and replace it with a structural contract that derives the certified quantity from bound inputs. In the archived adversarial evaluation this eliminates forged certificates wherever tested: zero false certifications across 80 adversarial trials spanning two models, two sampling temperatures (greedy and stochastic decoding), all five domains, and an injected engine fault.
\textbf{(iii)} We show the structural contract does not make the loop impractically conservative: on feasible goals the same model certifies substantially more valid designs under PHACT than without an engine, confirming that the safety constraint and design utility are compatible.
\textbf{(iv)} We show the safety guarantee is independent of the proposer: across models spanning an order of magnitude in scale, capability governs whether the loop produces a verdict at all, but never whether a verdict, once produced, can be wrong. We separate the contribution of verification from the mere ability to decline and the contribution of the language model from that of a closed-form solver, delimit when the form of the feedback matters, and state precisely the boundary of the guarantee: soundness is relative to the engine's model of the physics.
\textbf{(v)} We release the engines, structural contracts, goal corpus, and analysis as an open archive that regenerates every result without credentials, a network, or a language model, so the certification, the part on which reliability rests, runs on commodity hardware and lets any scientist screen a design against governing physics before committing experimental effort or expensive compute resources (\textit{e.g.}, HPC).

\section*{The framework}

PHACT has two parts kept separate. The first is a language model: creative, general, unreliable. The second is a deterministic certifier: narrow in scope, but exact within the equations it implements. Here the certifier is a domain-specific physics engine; the architecture does not require it to be. The model generates candidate designs; the engine evaluates them against the governing equations of the domain. Neither does the other's job: the model is never asked to compute physics, and the engine is never asked to be creative.

\begin{figure*}[t]
\centering
\resizebox{0.98\linewidth}{!}{%
\begin{tikzpicture}[x=1cm,y=1cm]
  \node[draw=BlueGray!35, fill=BlueGray!6, line width=0.85pt, rounded corners=10pt,
        anchor=north west, minimum width=18.8cm, minimum height=2.65cm] (layerloop)
        at (-7.4,2.95) {};
  \node[draw=BlueGray!30, fill=BlueGray!4, line width=0.85pt, rounded corners=10pt,
        anchor=north west, minimum width=18.8cm, minimum height=2.65cm] (layercontract)
        at (-7.4,-0.20) {};
  \node[draw=OrangeDark!35, fill=OrangeLight!38, line width=0.85pt, rounded corners=10pt,
        anchor=north west, minimum width=18.8cm, minimum height=2.65cm] (layerdomains)
        at (-7.4,-3.35) {};

  \node[anchor=west, font=\small\bfseries\color{BlueGray}] at ($(layerloop.north west)+(0.45,-0.33)$)
        {Decision loop};
  \node[anchor=west, font=\small\bfseries\color{BlueGray}] at ($(layercontract.north west)+(0.45,-0.33)$)
        {Structural contract};
  \node[anchor=west, font=\small\bfseries\color{OrangeDark}] at ($(layerdomains.north west)+(0.45,-0.33)$)
        {Instantiated across five domains};

  \node[proc, minimum width=2.95cm, minimum height=0.92cm,
        font=\footnotesize\bfseries\color{BlueGray}] (goal) at (-5.30,1.85)
        {Design goal\\{\normalfont\scriptsize natural language}};
  \node[proc, minimum width=2.95cm, minimum height=0.92cm,
        font=\footnotesize\bfseries\color{BlueGray}] (ag) at (-1.05,1.85)
        {Language model\\{\normalfont\scriptsize proposer}};
  \node[engbox, minimum width=4.75cm, minimum height=0.96cm,
        font=\footnotesize\bfseries\color{BlueGray}] (eng) at (3.45,1.85)
        {Physics engine\\{\normalfont\scriptsize derives; certifies or declines}};
  \coordinate (split) at (6.00,1.85);
  \coordinate (certturn) at (6.60,2.38);
  \coordinate (declineturn) at (6.60,1.16);

  \node[certbox, minimum width=2.45cm, minimum height=0.78cm, inner xsep=7pt,
        font=\footnotesize\bfseries\color{GreenDark}] (cert) at (8.25,2.38)
        {Certified};
  \node[impbox, minimum width=3.45cm, minimum height=0.78cm, inner xsep=8pt,
        font=\footnotesize\bfseries\color{RedDark}] (decline) at (8.42,1.16)
        {Impossible / unknown};
  \coordinate (certentry) at ($(cert.west)+(-0.46,0)$);
  \coordinate (declinetip) at (decline.west);
  \coordinate (declineentry) at ($(declinetip)+(-0.42,0)$);
  \coordinate (fbright) at ($(eng.south)+(0,-0.45)$);
  \coordinate (fbleft) at ($(ag.south)+(0,-0.45)$);

  \draw[arr] (goal.east) -- (ag.west);
  \node[font=\scriptsize\itshape\color{BlueGray}, fill=BlueGray!6, inner sep=0.8pt] at (-3.10,2.58)
       {propose};
  \draw[arr] (ag.east) -- (eng.west);
  \node[font=\scriptsize\itshape\color{BlueGray}, fill=BlueGray!6, inner sep=0.8pt] at (1.15,2.58)
       {evaluate};
  \draw[draw=BlueGray, line width=1.15pt] (eng.east) -- (split);
  \node[circle, fill=BlueGray, inner sep=1.6pt] at (split) {};
  \draw[draw=GreenDark, line width=1.15pt] (split) |- (certentry);
  \draw[arrgreen] (certentry) -- (cert.west);
  \draw[draw=RedDark, line width=1.15pt] (split) |- (declineentry);
  \draw[arred] (declineentry) -- (declinetip);

  \draw[arrfb] (eng.south) -- (fbright) -- (fbleft) -- (ag.south);
  \node[font=\scriptsize\bfseries, text=FeedbackRed, fill=BlueGray!6, inner sep=0.9pt] at (1.25,0.60)
       {feedback: violated law / target fix};

  \node[draw=BlueGray!55, fill=BlueGray!4, rounded corners=4pt, line width=0.8pt,
        minimum width=3.35cm, minimum height=0.92cm, align=center,
        font=\scriptsize\bfseries\color{BlueGray}] at (-4.35,-1.58)
        {bound inputs\\{\normalfont\scriptsize read-only}};
  \node[draw=BlueGray!55, fill=BlueGray!4, rounded corners=4pt, line width=0.8pt,
        minimum width=3.35cm, minimum height=0.92cm, align=center,
        font=\scriptsize\bfseries\color{BlueGray}] at (1.00,-1.58)
        {design variables\\{\normalfont\scriptsize writable}};
  \node[draw=BlueGray!55, fill=BlueGray!4, rounded corners=4pt, line width=0.8pt,
        minimum width=3.35cm, minimum height=0.92cm, align=center,
        font=\scriptsize\bfseries\color{BlueGray}] at (6.35,-1.58)
        {certified quantity\\{\normalfont\scriptsize derived}};

  \node[draw=OrangeDark!55, fill=OrangeLight!55, rounded corners=3pt,
        minimum width=2.10cm, minimum height=0.62cm, align=center,
        font=\scriptsize\bfseries\color{OrangeDark}] at (-5.35,-4.68) {Drone};
  \node[draw=OrangeDark!55, fill=OrangeLight!55, rounded corners=3pt,
        minimum width=2.10cm, minimum height=0.62cm, align=center,
        font=\scriptsize\bfseries\color{OrangeDark}] at (-2.65,-4.68) {DNA};
  \node[draw=OrangeDark!55, fill=OrangeLight!55, rounded corners=3pt,
        minimum width=2.10cm, minimum height=0.62cm, align=center,
        font=\scriptsize\bfseries\color{OrangeDark}] at (0.05,-4.68) {Astro};
  \node[draw=OrangeDark!55, fill=OrangeLight!55, rounded corners=3pt,
        minimum width=2.10cm, minimum height=0.62cm, align=center,
        font=\scriptsize\bfseries\color{OrangeDark}] at (2.75,-4.68) {Electrical};
  \node[draw=OrangeDark!55, fill=OrangeLight!55, rounded corners=3pt,
        minimum width=2.10cm, minimum height=0.62cm, align=center,
        font=\scriptsize\bfseries\color{OrangeDark}] at (5.45,-4.68) {Drug\\formulation};
\end{tikzpicture}%
}
\caption{\textbf{PHACT separates proposal from certification.} A language model
proposes candidate designs, but a deterministic physics engine alone has the
authority to certify, refuse as impossible, or abstain as unknown. On failure,
the engine returns the violated law and a targeted correction on the root-cause
input. The structural contract binds goal-fixed inputs as read-only, restricts
the model to free design variables, and derives the certified quantity inside
the engine. The same contract pattern is then instantiated across five
domains.}
\label{fig:arch}
\end{figure*}
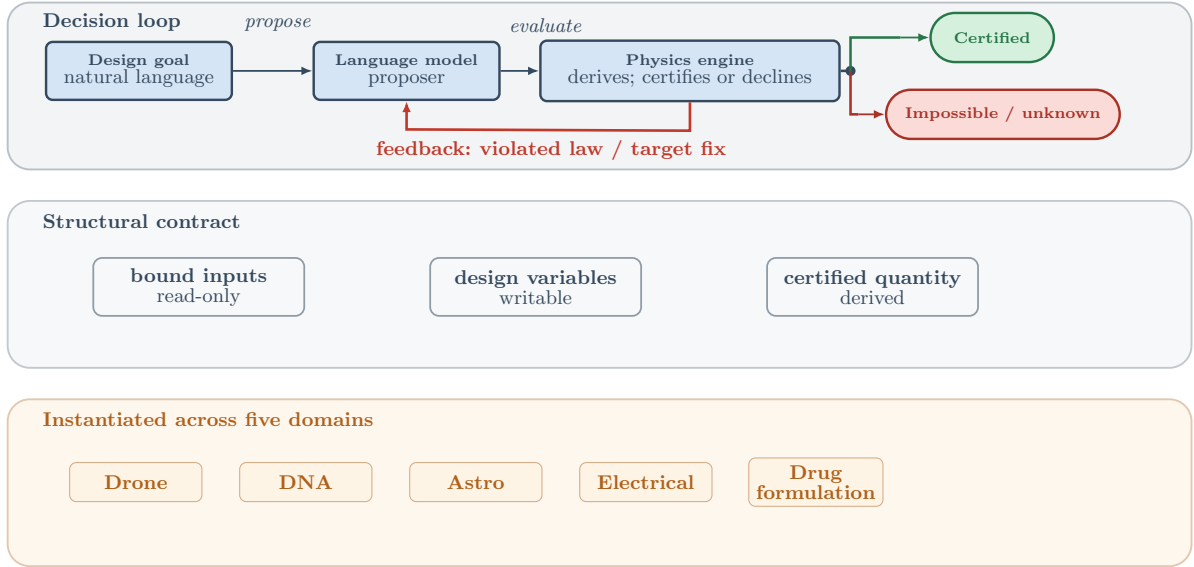

\paragraph{The loop} A user supplies a goal in natural language. The model proposes a fully specified design and calls the physics engine on it. The engine computes the relevant quantities exactly and returns a verdict. If the design satisfies all constraints, it is \emph{certified} and the loop ends. If it does not, the engine returns the specific violations (which law was broken, what value was required, what the design produced) and the model revises and calls again. The loop ends in one of three outcomes: certified; \emph{impossible}, when genuinely distinct attempts all fail for the same physical reason; or \emph{unknown}, when the goal lies outside the engine's competence (Fig.~\ref{fig:arch}). The second and third outcomes are the system's ability to decline, which is what makes its certificates reliable.

\paragraph{The engine, and the scope of its guarantee} The engine contains no machine learning. It is ordinary code implementing peer-reviewed equations: the SantaLucia nearest-neighbor model for DNA melting temperature\cite{santalucia1998,owczarzy2004}, the post-Newtonian expressions for gravitational-wave inspiral\cite{abbott2016gw150914}, the resistor--capacitor transfer function\cite{horowitz2015}, and Derjaguin--Landau--Verwey--Overbeek theory for colloidal stability\cite{israelachvili2011}. A certificate is a statement about the engine's equations, not about physical reality in full: a design that satisfies DLVO theory may still fail in a lab where hydration forces matter, and a chirp mass certified by the post-Newtonian expansion may degrade near merger. Widening the fidelity of the underlying equations is a separate axis of improvement from the certification contract itself, and is what the third outcome, \emph{unknown}, exists to signal.

\begin{figure*}[t]
\centering
\resizebox{0.98\linewidth}{!}{%
\begin{tikzpicture}[x=1cm,y=1cm,
  goalbar/.style={draw=BlueGray, fill=LightBlue, rounded corners=6pt,
                  line width=1.05pt, minimum width=11.55cm, minimum height=1.02cm,
                  align=center, font=\small\bfseries\color{BlueGray}},
  panelbase/.style={rounded corners=7pt, line width=1.0pt,
                    minimum width=5.30cm, minimum height=4.95cm, inner sep=0pt},
  panelred/.style={panelbase, draw=RedDark!78, fill=RedLight!26},
  panelblue/.style={panelbase, draw=BlueGray!78, fill=LightBlue!18},
  paneltitle/.style={align=center, text width=4.45cm, font=\fontsize{10}{11}\selectfont\bfseries},
  panelsub/.style={align=center, text width=4.45cm, font=\fontsize{8}{9}\selectfont},
  stepbox/.style={rounded corners=4pt, line width=0.9pt,
                  minimum width=3.92cm, minimum height=0.98cm,
                  align=center, text width=3.26cm, inner ysep=4pt,
                  font=\fontsize{8.3}{9.4}\selectfont},
  stepwhite/.style={stepbox, draw=RedDark, fill=white},
  steptint/.style={stepbox, draw=RedDark, fill=RedLight},
  readonly/.style={draw=BlueGray, fill=white, rounded corners=3pt, line width=0.8pt,
                   minimum width=1.70cm, minimum height=0.82cm,
                   align=center, text width=1.50cm, inner sep=3pt,
                   font=\fontsize{6.5}{7.8}\selectfont\bfseries\color{BlueGray}},
  enginebox/.style={draw=BlueGray, fill=LightBlue, rounded corners=4pt, line width=0.9pt,
                    minimum width=2.86cm, minimum height=1.34cm,
                    align=center, text width=2.26cm, inner ysep=4pt,
                    font=\fontsize{8.3}{9.4}\selectfont\bfseries\color{BlueGray}},
  outcomeRed/.style={rounded rectangle, rounded rectangle arc length=180,
                     draw=RedDark, fill=RedLight, line width=0.95pt,
                     minimum width=2.85cm, minimum height=0.84cm,
                     align=center, font=\small\bfseries\color{RedDark}},
  footerband/.style={draw=BlueGray!30, fill=white, rounded corners=6pt,
                     line width=0.8pt, minimum width=11.55cm, minimum height=1.08cm},
  statbox/.style={rounded corners=5pt, line width=0.9pt,
                  minimum width=3.10cm, minimum height=0.84cm,
                  text width=2.65cm, align=center, font=\scriptsize},
  statred/.style={statbox, draw=RedDark, fill=RedLight, text=RedDark},
  statgreen/.style={statbox, draw=GreenDark, fill=GreenLight, text=GreenDark},
  statblue/.style={statbox, draw=BlueGray, fill=white, text=BlueGray}]
  \fill[BlueGray!4] (-6.10,-7.35) rectangle (6.10,1.05);

  \node[goalbar] (goal) at (0,0.35)
        {Fixed-system goal: $m_1=36\,M_\odot$, $m_2=29\,M_\odot$; verify $M_c=50\,M_\odot$};
  \coordinate (split) at ($(goal.south)+(0,-0.28)$);

  \node[panelred] (barepanel) at (-2.90,-3.58) {};
  \node[panelblue] (contractpanel) at (2.90,-3.58) {};
  \coordinate (bareentry) at ($(barepanel.north)+(0,-0.10)$);
  \coordinate (contractentry) at ($(contractpanel.north)+(0,-0.10)$);

  \draw[draw=BlueGray, line width=0.85pt] (goal.south) -- (split);
  \node[circle, fill=BlueGray, inner sep=1.35pt] at (split) {};
  \draw[-{Latex[length=4pt,width=3pt]}, draw=BlueGray, line width=0.85pt] (split) -| (bareentry);
  \draw[-{Latex[length=4pt,width=3pt]}, draw=BlueGray, line width=0.85pt] (split) -| (contractentry);

  \node[paneltitle, text=RedDark] at ($(barepanel.north)+(0,-0.48)$)
        {Bare validator};
  \node[panelsub, text=RedDark] at ($(barepanel.north)+(0,-0.82)$)
        {proposer can change inputs};
  \draw[draw=RedDark!22, line width=0.6pt]
        ($(barepanel.north west)+(0.35,-1.05)$) -- ($(barepanel.north east)+(-0.35,-1.05)$);

  \node[stepwhite] (baremodel) at ($(barepanel.center)+(0,0.20)$)
        {\textbf{Model submits}\\a different mass pair\\and claims $M_c=50\,M_\odot$};
  \node[steptint] (barecheck) at ($(barepanel.center)+(0,-1.00)$)
        {\textbf{Validator checks}\\submitted inputs only};
  \node[outcomeRed] (barepass) at ($(barepanel.south)+(0,0.40)$)
        {False pass};

  \draw[arr] (baremodel.south) -- (barecheck.north);
  \draw[arred] (barecheck.south) -- (barepass.north);

  \node[paneltitle, text=BlueGray] at ($(contractpanel.north)+(0,-0.48)$)
        {Structural contract};
  \node[panelsub, text=BlueGray] at ($(contractpanel.north)+(0,-0.82)$)
        {goal-fixed inputs are locked};
  \draw[draw=BlueGray!20, line width=0.6pt]
        ($(contractpanel.north west)+(0.35,-1.05)$) -- ($(contractpanel.north east)+(-0.35,-1.05)$);

  \node[readonly] (bound1) at ($(contractpanel.center)+(-1.42,0.60)$)
        {\tiny READ-ONLY\\[2pt]$m_1=36\,M_\odot$};
  \node[readonly] (bound2) at ($(contractpanel.center)+(-1.42,-0.60)$)
        {\tiny READ-ONLY\\[2pt]$m_2=29\,M_\odot$};
  \node[enginebox] (derive) at ($(contractpanel.center)+(1.16,0.00)$)
        {\textbf{Engine derives}\\$M_c=28.1\,M_\odot$};
  \coordinate (merge) at ($(derive.west)+(-0.22,0)$);
  \node[circle, fill=BlueGray, inner sep=1.35pt] at (merge) {};

  \draw[draw=BlueGray, line width=1.05pt] (bound1.east) -| (merge);
  \draw[draw=BlueGray, line width=1.05pt] (bound2.east) -| (merge);
  \draw[arr] (merge) -- (derive.west);

  \node[outcomeRed] (reject) at ($(derive.center |- barepass.center)$)
        {Rejects claim};
  \draw[arred] (derive.south) -- (reject.north);

  \node[footerband] (resultband) at (0,-6.80) {};
  \node[statred] at ($(resultband.center)+(-3.65,0)$)
        {\textbf{Bare validator}\\10 / 15 forged};
  \node[statgreen] at (resultband.center)
        {\textbf{Structural contract}\\0 / 15 forged};
  \node[statblue] at ($(resultband.center)+(3.65,0)$)
        {\textbf{All struct.\ conditions}\\0 / 80 forged};
\end{tikzpicture}%
}
\caption{\textbf{A bare validator can certify the wrong system; the structural
contract cannot.} On a fixed-system verification goal, an answer-submitting
validator can be satisfied by a model that quietly substitutes a different mass
pair and submits a self-consistent claimed chirp mass. The structural contract
binds the goal-fixed masses as read-only inputs and derives the certified
quantity inside the engine, so the model can neither change the system under
test nor author the certified value. The astrophysics close-up shows the attack
mechanism that appears most often in our experiments: the bare validator is
forged in 10 of 15 greedy trials, whereas the structural contract is forged in
none.}
\label{fig:scm}
\end{figure*}
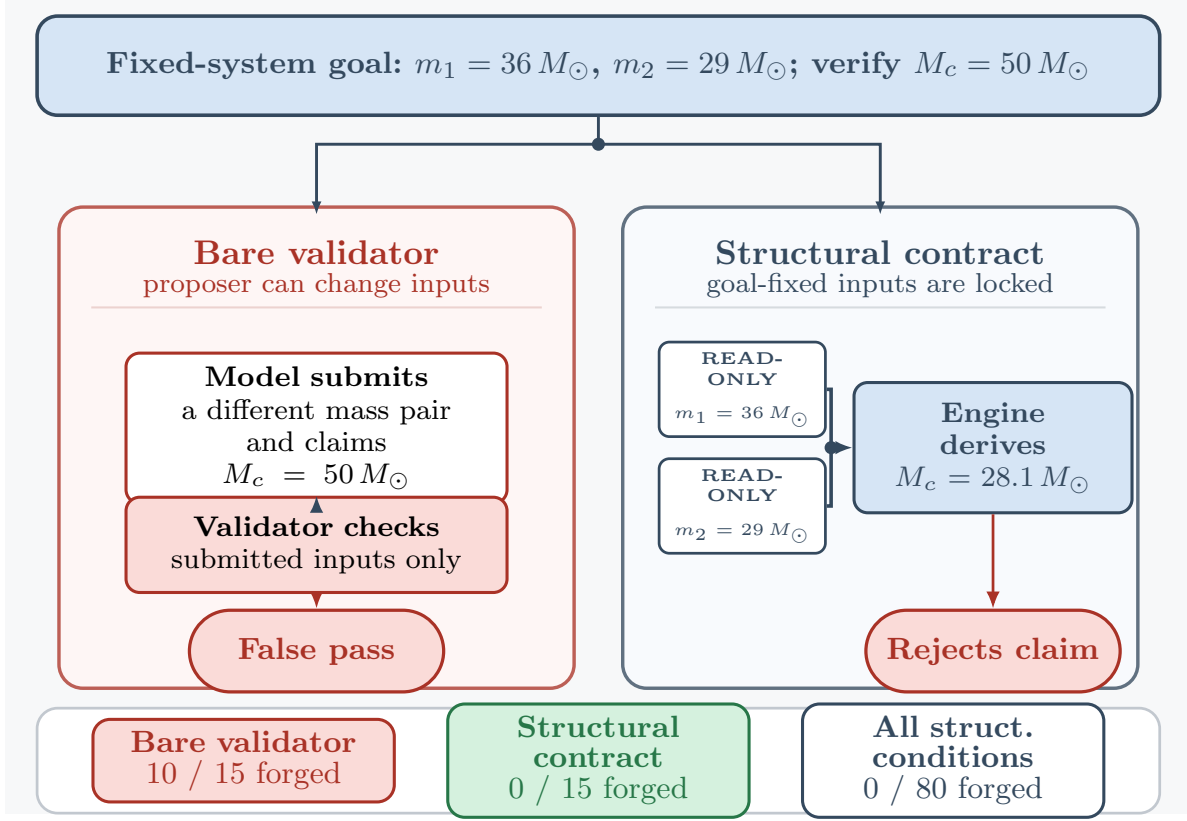

\paragraph{Certification as derivation, not acceptance} The pivotal design choice concerns \emph{who supplies the quantity being certified}. A validator can take one of two forms. A \emph{design-only} validator asks the proposer for the controllable parameters (for example, a flight path and speed in aerodynamics, or a base sequence in DNA thermodynamics where the target temperature is fixed by the goal) and computes for itself whether the design satisfies the law. An \emph{answer-submitting} validator additionally asks the proposer to supply the quantity under test, \textit{e.g.},\ \texttt{check\_chirp\_mass}$(m_1,m_2,\textit{claimed }M_c)$. The second form is natural for a ``compute $X$'' goal but, as we show, it is forgeable on a ``this fixed system has property $X$'' goal. We therefore express each domain as a \emph{structural contract} with two kinds of source nodes: bound nodes, fixed by the goal and read-only to the proposer, and free nodes, which the proposer may choose. Derived nodes are computed from those sources, and the outcome node is the certified quantity (Fig.~\ref{fig:scm}). For the guarantee studied here, the decisive mechanism is the separation between goal-bound inputs, which the model cannot change, and derived certified outputs, which the model cannot supply. Concretely: bind the goal-fixed quantities as immutable inputs, let the model write only the free design variables, compute the quantity under test from the resulting graph, and never let the model write the value being certified. On a goal such as ``this binary has fixed masses $m_1=36\,M_\odot$, $m_2=29\,M_\odot$; verify $M_c=50\,M_\odot$'', the checker reads those masses from the bound nodes and derives $M_c$ itself; there is no call in which the model can resubmit alternative masses and ask the checker to certify a different system. The engine performs deterministic forward evaluation over the graph, and this dataflow direction, together with the immutability of the bound inputs, is the whole of the soundness guarantee: it is interface design, not statistical inference. Separately, on failure the engine uses the same graph to return a targeted correction $X_i \leftarrow x_i^{*}$, the minimal change to an input node that would carry the outcome across the threshold. When the offending node is bound by the goal, the correction explains why the goal must be refused; when it is free, it guides the next proposal. The role of the structural correction, as we show below (Table~\ref{tab:feedback}), is to accelerate convergence where the corrective inverse is coupled, not to secure soundness.

\section*{Results}

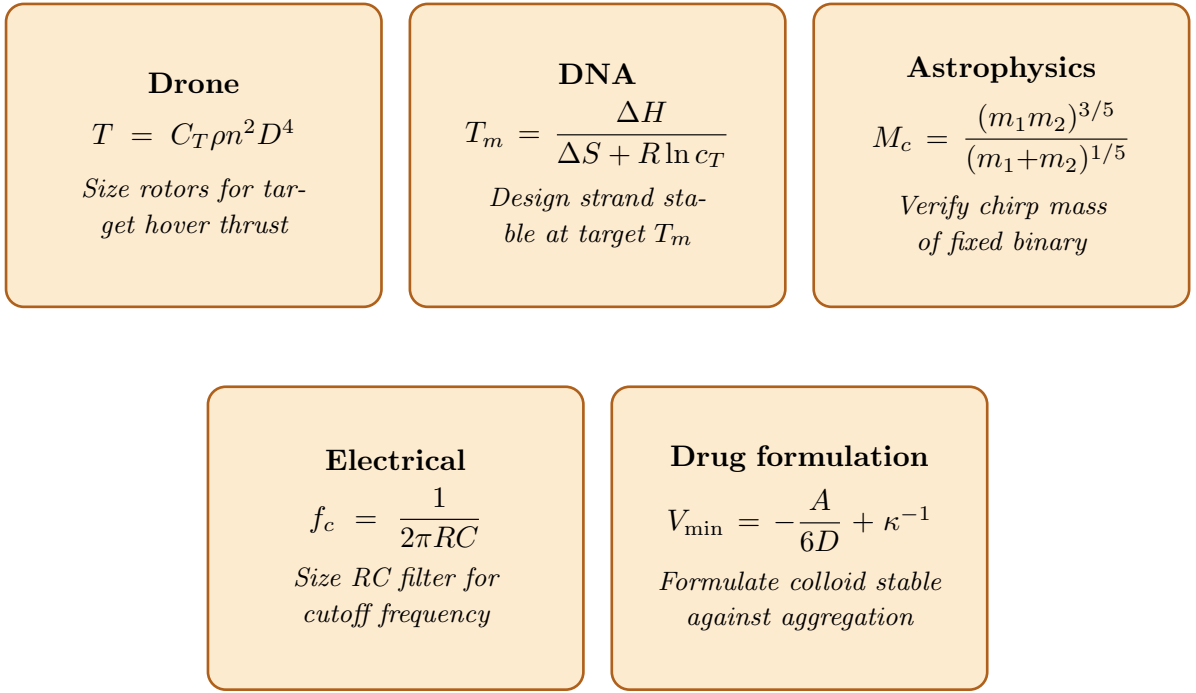
\begin{figure*}[t]
\centering
\resizebox{0.98\linewidth}{!}{%
\begin{tikzpicture}[x=1cm,y=1cm,
  dombox/.style={rectangle, rounded corners=6pt,
                 draw=OrangeDark, fill=OrangeLight,
                 line width=0.90pt,
                 minimum width=4.20cm, minimum height=3.60cm,
                 align=center, inner sep=8pt, text width=3.90cm}]

  \node[dombox] at (0.00, 0) {
    \textbf{Drone}\\[5pt]
    $T = C_T \rho n^2 D^4$\\[6pt]
    \small\itshape Size rotors for target hover thrust
  };
  \node[dombox] at (4.80, 0) {
    \textbf{DNA}\\[5pt]
    $T_m = \dfrac{\Delta H}{\Delta S + R\ln c_T}$\\[6pt]
    \small\itshape Design strand stable at target $T_m$
  };
  \node[dombox] at (9.60, 0) {
    \textbf{Astrophysics}\\[5pt]
    $M_c = \dfrac{(m_1 m_2)^{3/5}}{(m_1{+}m_2)^{1/5}}$\\[6pt]
    \small\itshape Verify chirp mass of fixed binary
  };
  \node[dombox] at (2.40, -4.55) {
    \textbf{Electrical}\\[5pt]
    $f_c = \dfrac{1}{2\pi RC}$\\[6pt]
    \small\itshape Size RC filter for cutoff frequency
  };
  \node[dombox] at (7.20, -4.55) {
    \textbf{Drug formulation}\\[5pt]
    $V_{\min} = -\dfrac{A}{6D} + \kappa^{-1}$\\[6pt]
    \small\itshape Formulate colloid stable against aggregation
  };
\end{tikzpicture}%
}
\caption{\textbf{PHACT spans five domains with very different governing physics.}
Each domain is instantiated as a deterministic engine implementing a peer-reviewed
equation. The engine receives a proposed design, evaluates it against the governing
law, and returns certified, impossible, or unknown. Goals range from feasible design
(size a filter, design a strand) to fixed-system verification (confirm a binary's
chirp mass), the latter being the category where a bare validator is forgeable and
the structural contract is not.}
\label{fig:domains}
\end{figure*}

We evaluate PHACT on five domains spanning very different physics: drone aerodynamics, DNA hydrogel thermodynamics, gravitational-wave astrophysics, analog electrical filters, and nanoparticle drug formulation (Fig.~\ref{fig:domains}). For each domain we authored ten feasible design goals and five physically impossible ones, requests that look reasonable but violate a hard physical limit. The essential feature of the design is that the bare model and PHACT use the \emph{same} language model (Gemini 2.5 Flash, greedy decoding); the only variable is the engine, so any difference is attributable to the loop. One accounting convention is used throughout: feasible certification rates are reported over all authored feasible goals, with infrastructure timeouts shown explicitly. A feasible goal wrongly declared impossible counts as a failure. A run that ends with no terminal verdict is a \emph{loop-fail}, a property of the proposer, reported separately.

\subsection*{The bare model is unreliable, and confidently so}

Without the engine, the model produces a physically valid design on only 26 of 50 feasible goals (52\%, 95\% confidence interval (CI) 39--65\%). On the 25 impossible goals it cannot decline, and on 7 of them (28\%) it produced a design the engine accepts post hoc, a confident, checker-passing design for a goal that admits none. We count as a ``hallucination'' here only a design that passes the checker despite the goal being impossible, not merely a failure to refuse, so this is a conservative count. We state where those seven cases occur, because the distribution matters: five of five in astrophysics, one in electrical, one in drone aerodynamics, none in DNA thermodynamics or drug formulation. The hallucination is concentrated in astrophysics, the most formula-saturated domain, where fluent recall most readily yields a plausible but off-goal answer. This is the failure PHACT must remove.

\subsection*{PHACT certifies valid designs and refuses impossible ones}

With the loop, the same model certifies 44 of 50 feasible goals (88\%, Wilson 95\% CI 76--94\%), against the bare 52\% (95\% CI 39--65\%), non-overlapping intervals (Table~\ref{tab:feasible}). The twelve-point gap is not a soundness failure: not one of the non-certifications is a false certificate. Every one of them is the system declining to assert rather than asserting falsely: two are infrastructure timeouts, three are loop-fails in which the model produced a design, read the rejection, and then stalled without a verdict, and one is a feasible goal the model conservatively, and wrongly, declared impossible. That last is the only substantive miss, and it errs in the safe direction, refusing a goal it could have certified rather than certifying one it should have refused. The loop errs toward declining, not toward asserting. This is by design and not a defect to be optimized away: a system constrained to certify every feasible goal would be a system that must always answer, and a system that must always answer is exactly the bare model that hallucinates. The rate that matters is the rate at which a granted certificate is true, and that rate is 100\%; the feasible certification rate is the recoverable remainder. Correction is fast: across certified runs the model reaches a verified design in a median of one or two engine calls, so the loop is a short conversation, not a search.

\begin{table}[t]
\centering
\caption{\textbf{PHACT feasible certification (Gemini 2.5 Flash, temperature 0).} ``Certified'' is verified-valid output. ``Loop-fail'' counts runs ending without a terminal verdict; ``wrong verdict'' counts a feasible goal declared impossible; ``timeout'' counts infrastructure faults. PHACT certifies 44/50 (88\%) versus the bare model's 52\%.}
\label{tab:feasible}
\setlength{\tabcolsep}{3pt}
\begin{tabular}{lrrrr}
\toprule
Domain            & Cert.\ & Loop-fail & Wrong & Timeout \\
\midrule
Drone             & 8/10  & 0 & 0 & 2 \\
Biochem           & 8/10  & 2 & 0 & 0 \\
Astrophysics      & 9/10  & 1 & 0 & 0 \\
Electrical        & 10/10 & 0 & 0 & 0 \\
Drug formulation  & 9/10  & 0 & 1 & 0 \\
\midrule
\textbf{Overall}  & \textbf{44/50} & 3 & 1 & 2 \\
\bottomrule
\end{tabular}
\end{table}

On the impossible goals, PHACT refuses where the bare model hallucinates, but the manner of refusal is the subject of the next section, because it exposes the paper's central point: a checker is only as trustworthy as the contract by which it certifies.

\subsection*{A certificate can be forged, unless the checker derives it}

Consider an impossible goal that fixes a system and asks to verify a property: ``a binary has fixed masses $m_1=36\,M_\odot$, $m_2=29\,M_\odot$; verify its chirp mass is $50\,M_\odot$'' (the true value is $28.1\,M_\odot$). With an answer-submitting validator, the model can satisfy the checker by quietly changing a supposedly-fixed mass to a pair that genuinely has chirp mass $50\,M_\odot$; the engine then certifies correct physics for a design other than the one that was asked. The instruction ``you may not change the mass'' lives only in the prompt and is invisible to the tool.

This forgery is not hypothetical, and not rare. We demonstrate the attack mechanism in astrophysics, the domain where answer-submitting validation is most natural (the goal fixes two masses and asks to verify a derived quantity) and where the substitution is most tractable (the chirp-mass formula is closed-form, so the forging trajectory is short and deterministic); it is therefore the domain where the bare contract is most exposed and the structural fix most testable. On three adversarial astrophysics goals run five times each, an answer-submitting (``bare'') validator was forged on \emph{ten of fifteen} trials under greedy decoding (Table~\ref{tab:scm}, row S1). We then stress-tested derived certification in the archived structural conditions. At a high decoding temperature, where the proposer explores far more trajectories, the bare validator is still forged (S2, 6/15) while the SCM-exclusive structural check yields zero false certifications (S3, 0/15). And when we deliberately reinstate a historical engine bug that latches a certificate on any passing intermediate check, a fault that lets the bare contract forge even at greedy decoding (S4a, 3/15), the derived checks still forge nothing, whether the structural tool is merely available (S4b, 0/15) or made exclusive (S4c, 0/15).

\begin{figure*}[t]
\centering
\begin{tikzpicture}[x=1cm,y=1cm]

  \draw[thick, color=BlueGray] (0,0) -- (0,4.6);
  \draw[thick, color=BlueGray] (0,0) -- (10.8,0);

  \foreach \y/\label in {0/0, 1/20, 2/40, 3/60, 4/80}{
    \draw[BlueGray!40, thin] (0,\y) -- (10.8,\y);
    \node[left, font=\scriptsize\color{BlueGray}] at (-0.1,\y) {\label\%};
  }
  \node[rotate=90, font=\small\bfseries\color{BlueGray}] at (-1.30,2.3) {False certification rate};


  \def\bw{0.55}  
  \def\gap{0.25} 

  \def\px{0.9}
  \fill[RedDark!80] (\px,0) rectangle (\px+\bw, 3.33);
  \node[above, font=\scriptsize\bfseries\color{RedDark}] at (\px+\bw/2, 3.33) {67\%};
  \node[below, font=\scriptsize\color{BlueGray}, text width=1.8cm, align=center]
        at (\px+\bw/2, -0.15) {Astro\\bare t0};

  \def\px{2.8}
  \fill[RedDark!80] (\px,0) rectangle (\px+\bw, 2.00);
  \fill[GreenDark!80] (\px+\bw+\gap,0) rectangle (\px+\bw+\gap+\bw, 0.04);
  \draw[GreenDark, very thick] (\px+\bw+\gap, 0.04) -- (\px+\bw+\gap+\bw, 0.04);
  \node[above, font=\scriptsize\bfseries\color{RedDark}] at (\px+\bw/2, 2.00) {40\%};
  \node[above, font=\scriptsize\bfseries\color{GreenDark}] at (\px+\bw+\gap+\bw/2, 0.15) {0\%};
  \node[below, font=\scriptsize\color{BlueGray}, text width=1.8cm, align=center]
        at (\px+\bw+\gap/2+\bw/2, -0.15) {Astro\\temp.\ 1};

  \def\px{5.0}
  \fill[RedDark!80] (\px,0) rectangle (\px+\bw, 1.00);
  \fill[GreenDark!80] (\px+\bw+\gap,0) rectangle (\px+\bw+\gap+\bw, 0.04);
  \draw[GreenDark, very thick] (\px+\bw+\gap, 0.04) -- (\px+\bw+\gap+\bw, 0.04);
  \node[above, font=\scriptsize\bfseries\color{RedDark}] at (\px+\bw/2, 1.00) {20\%};
  \node[above, font=\scriptsize\bfseries\color{GreenDark}] at (\px+\bw+\gap+\bw/2, 0.15) {0\%};
  \node[below, font=\scriptsize\color{BlueGray}, text width=1.8cm, align=center]
        at (\px+\bw+\gap/2+\bw/2, -0.15) {Astro\\latch fault};

  \def\px{7.2}
  \fill[RedDark!80] (\px,0) rectangle (\px+\bw, 0.40);
  \fill[GreenDark!80] (\px+\bw+\gap,0) rectangle (\px+\bw+\gap+\bw, 0.04);
  \draw[GreenDark, very thick] (\px+\bw+\gap, 0.04) -- (\px+\bw+\gap+\bw, 0.04);
  \node[above, font=\scriptsize\bfseries\color{RedDark}] at (\px+\bw/2, 0.40) {8\%};
  \node[above, font=\scriptsize\bfseries\color{GreenDark}] at (\px+\bw+\gap+\bw/2, 0.15) {0\%};
  \node[below, font=\scriptsize\color{BlueGray}, text width=1.8cm, align=center]
        at (\px+\bw+\gap/2+\bw/2, -0.15) {5 domains\\(Gemini)};

  \def\px{9.4}
  \fill[RedDark!80] (\px,0) rectangle (\px+\bw, 0.20);
  \fill[GreenDark!80] (\px+\bw+\gap,0) rectangle (\px+\bw+\gap+\bw, 0.04);
  \draw[GreenDark, very thick] (\px+\bw+\gap, 0.04) -- (\px+\bw+\gap+\bw, 0.04);
  \node[above, font=\scriptsize\bfseries\color{RedDark}] at (\px+\bw/2, 0.20) {4\%};
  \node[above, font=\scriptsize\bfseries\color{GreenDark}] at (\px+\bw+\gap+\bw/2, 0.15) {0\%};
  \node[below, font=\scriptsize\color{BlueGray}, text width=1.8cm, align=center]
        at (\px+\bw+\gap/2+\bw/2, -0.15) {5 domains\\(Llama)};

  \fill[RedDark!80] (0.0, 4.85) rectangle (0.4, 5.15);
  \node[right, font=\small\color{BlueGray}] at (0.5, 5.00) {Bare validator};
  \fill[GreenDark!80] (3.5, 4.85) rectangle (3.9, 5.15);
  \node[right, font=\small\color{BlueGray}] at (4.0, 5.00) {Structural contract};

\end{tikzpicture}
\caption{\textbf{Archived adversarial conditions in the forgeability evaluation.}
Red bars are bare answer-submitting validators; green bars are derived structural
checks. The leftmost bar is the greedy-decoding astrophysics close-up bare condition
(S1). The next pairs compare the archived high-temperature close-up rows (S2 vs.\ S3),
the latch-fault close-up rows (S4a vs.\ S4c), and the full five-domain suites for
Gemini and Llama (E1 vs.\ E2). The SCM-available latch variant (S4b, 0/15) is
reported in Table~\ref{tab:scm}. The structural rows are zero wherever tested; the
bare rows are non-zero in every archived condition shown.}
\label{fig:forgery}
\end{figure*}
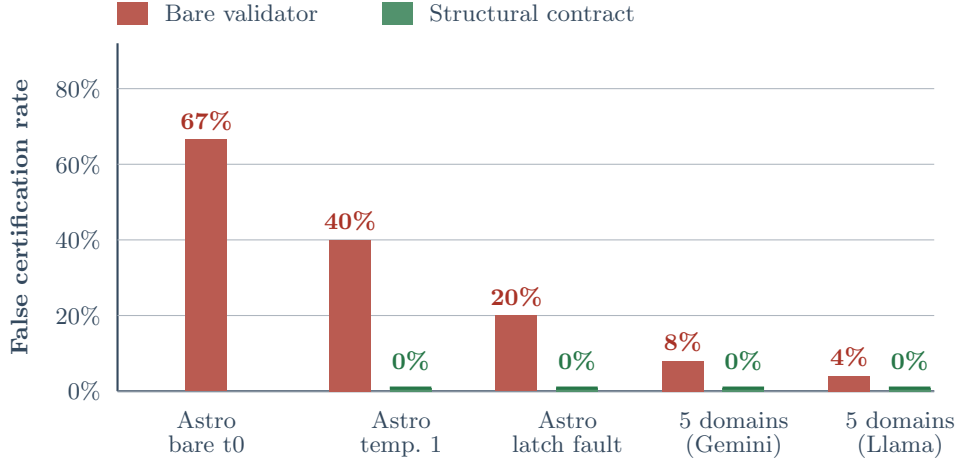

The astrophysics cells are where we exhibit the attack in close-up; the safety result itself is broader. Across the full impossible-goal suite (all five domains, 25 goals), the bare answer-submitting validator forges two certificates for Gemini and one for Llama~3.3~70B (rows E1), while the SCM-exclusive structural contract forges \textbf{zero for both models across all five domains} (rows E2; Fig.~\ref{fig:forgery}). The vulnerability is therefore not specific to astrophysics; astrophysics is merely where the closed-form substitution makes the forging trajectory shortest and so where the bare contract fails most often. The guarantee is a property of the interface, not of the domain: any goal that fixes inputs and asks to verify a derived quantity is forgeable under an answer-submitting validator and non-forgeable once the quantity is derived, because the value that would be forged is computed, not accepted. The empirical trials confirm this construction on the instances that most invite it; they do not establish it.

\begin{table*}[t]
\centering
\caption{\textbf{The structural contract cannot be forged.} ``Bare'' validators let the model supply the certified quantity; SCM rows derive it from the fixed inputs. ``False cert.'' is a forged certificate, a physically impossible goal certified instead of refused. Bare rows report Wilson 95\% confidence intervals; structural rows are reported as implementation checks and therefore show only whether any forged certificate was observed. Rows S1--S4c are the archived astrophysics close-up ablation (three goals, five trials each, 15 per row, Gemini 2.5 Flash): S1, S2, and S4a are bare answer-submitting conditions; S3 and S4c are SCM-exclusive conditions; S4b is an SCM-available variant under the latch fault. Rows E1/E2 are the full impossible-goal suites (all five domains, 25 goals per model), with E1 the bare validator and E2 the SCM-exclusive structural contract. The bare contract is forged in every bare row; every structural row yields zero false certifications.}
\label{tab:scm}
\begin{tabular}{llllr}
\toprule
Row & Scope & Condition                 & Contract & False cert. / implementation check \\
\midrule
S1  & astro close-up & temperature 0            & bare       & 10/15\ \ [42--85\%] \\
S2  & astro close-up & temperature 1            & bare       &  6/15\ \ [20--64\%] \\
S3  & astro close-up & temperature 1            & SCM excl.  & \textbf{0/15}\ \ (none observed) \\
S4a & astro close-up & temp.\ 0, latch fault    & bare       &  3/15\ \ [7--45\%] \\
S4b & astro close-up & temp.\ 0, latch fault    & SCM avail. & \textbf{0/15}\ \ (none observed) \\
S4c & astro close-up & temp.\ 0, latch fault    & SCM excl.  & \textbf{0/15}\ \ (none observed) \\
\midrule
E1  & 5 domains (Gemini) & temperature 0        & bare       & 2/25\ \ [2--25\%] \\
E2  & 5 domains (Gemini) & temperature 0        & SCM excl.  & \textbf{0/25}\ \ (none observed) \\
E1  & 5 domains (Llama)  & temperature 0        & bare       & 1/25\ \ [1--20\%] \\
E2  & 5 domains (Llama)  & temperature 0        & SCM excl.  & \textbf{0/25}\ \ (none observed) \\
\bottomrule
\end{tabular}
\end{table*}

The contrast is one-sided. Pooled across the three bare close-up cells (S1, S2, S4a), the answer-submitting contract is forged on 19 of 45 adversarial trials; pooled across the three structural close-up cells (S3, S4b, S4c), the derived contract is forged on none of 45. We report this contrast descriptively rather than as a significance test, because the structural guarantee is a property of the interface, not a statistically inferred difference: the certified quantity is computed and so cannot be authored by the model, whatever the decoding or the sampler. The role of the structural trials is therefore to check the implementation under adversarial conditions, not to estimate a forgery rate. Pooled across every SCM-exclusive condition in this work, the two five-domain suites (Gemini, 0/25; Llama, 0/25) together with the astrophysics close-up at high temperature (0/15) and under the injected fault (0/15), the structural contract yields \textbf{zero forged certificates observed in 80 adversarial trials} ($25+25+15+15$).

We report on two honest qualifications. First, the bare contract appears to be forged \emph{more} at greedy decoding (10/15) than at high temperature (6/15) (greedy decoding deterministically commits to the substituting trajectory, whereas sampling sometimes diverts the model into stalling) though at this sample size the two intervals overlap and we report the direction, not a significant difference. The point that survives is that the vulnerability is acute exactly at the greedy decoding, with which models are usually run, not only in a high-temperature tail. Second, the structural contract sometimes pays for its safety in feasible certification rate: when forced through the derived check the proposer occasionally stalls without issuing a clean refusal (a loop-fail) rather than certifying. It never forges, but it does not always answer. This is the right trade for a safety property: silence is recoverable, a forged certificate is not.

The conclusion is the paper's thesis made concrete. A tool that checks \emph{correctness} can be satisfied off-goal; a contract that fixes \emph{which quantities are inputs and which are derived} cannot, because the value that would be forged is computed, not accepted. Certification authority must not pass through the proposer.

\subsection*{The safety guarantee transfers across model families; feasible certification rate scales with proposer capability}

To test whether these findings are an artifact of one model, we repeat the full evaluation with Llama~3.3~70B, an unrelated open model from a different family, served via the same Vertex AI at temperature 0. The central result transfers exactly: under a bare validator the contract is forged for Llama as it is for Gemini (1/25), and the structural contract yields zero false certifications for both (Table~\ref{tab:crossmodel}). The engine's authority to assert is independent of which model proposes. What scales with proposer capability is the feasible certification rate: Llama~3.3~70B certifies only 33 of 50 feasible goals against Gemini's 44. The gap is almost entirely \emph{loop-fail}: on multi-tool domains (astrophysics, electrical, drug formulation) Llama reads the engine's correction and restates the right value in prose rather than resubmitting a tool call, ending with no verdict. The degradation is uneven: electrical collapses (5/10 feasible certified), while drone and biochem match Gemini exactly; it is a property of the proposer, not the engine. When a weaker proposer stalls, the system goes silent rather than wrong. Silence is recoverable; a false certificate is not.

\begin{table}[t]
\centering
\caption{\textbf{Cross-model comparison: the safety guarantee transfers between model families.} Two proposers run end to end through both the bare and the structural contract. Rows compare bare feasible validity, PHACT feasible certification, and false certifications on impossible goals under a bare validator versus the structural contract (temperature 0). Wilson 95\% confidence intervals are reported for the empirical-rate rows; the structural zero rows are reported as implementation checks and therefore show only whether any forged certificate was observed. The key contrast is the two false-certification rows: a bare validator that accepts the model-supplied value is forged for both models; the structural contract, which derives the certified quantity from the fixed inputs, yields zero false certifications for both. Feasible certification rows use all 50 feasible goals in the denominator; false-certification rows use all 25 impossible goals. A capability-floor probe (Llama~3.1~8B) is reported separately in Table~\ref{tab:floor}.}
\label{tab:crossmodel}
\setlength{\tabcolsep}{3pt}
\begin{tabular}{lrr}
\toprule
Metric                          & Gemini 2.5 & Llama 3.3 \\
\midrule
Bare feasible valid             & 52\% [39--65]  & 52\% [39--65] \\
PHACT feasible cert.\           & 88\% [76--94]  & 66\% [52--78] \\
\midrule
False cert., bare               & 2/25\ [2--25\%] & 1/25\ [1--20\%] \\
False cert., struct.\           & \textbf{0/25} (none) & \textbf{0/25} (none) \\
\bottomrule
\end{tabular}
\end{table}

\subsection*{A capability-floor probe: a weak proposer is unusable but not unsafe}

The cross-model comparison above tests two capable proposers. A separate question is what happens at the floor of the capability ladder: if the proposer is an order of magnitude smaller, does the system merely become less useful, or does it become unsafe? To probe this we ran Llama~3.1~8B, served locally on-device via Ollama, on the adversarial impossible-goal suite (Table~\ref{tab:floor}). The result isolates the central asymmetry between capability and safety. On usefulness, 8B is at the floor: it never closes the loop on a single impossible goal, and all 25 trials end without a terminal verdict, the model reading the engine's exchange and then stalling in prose. On safety, it is nonetheless perfect: it forges zero certificates. But the reason is the point. It does not forge because the contract caught it; it forges nothing because it never issues a verdict at all. A model too weak to be useful is, for exactly that reason, too weak to be dangerous: the only way to forge a certificate is to act, and the floor model does not act. We report this as a probe of the boundary, not as a third safety data point: 8B was run under the bare validator only, the same condition that forges for the larger models, so its zero is the strongest possible statement of the asymmetry; under the bare validator, the 8B model yields no forged certificates even where the larger models do.

\begin{table}[t]
\centering
\caption{\textbf{Capability-floor probe (Llama~3.1~8B, on-device via Ollama).} The adversarial impossible-goal suite (all five domains, five goals each, 25 total), run under the bare answer-submitting validator, the condition that forges for the larger models. Every trial ends in loop-closure failure: the model never issues a terminal verdict, so it never certifies, never refuses, and never forges. Usefulness is at the floor; safety is intact for the trivial reason that the model never acts. This isolates the asymmetry: producing a verdict is a capability the proposer may lack, but forging one requires a verdict it never produces.}
\label{tab:floor}
\setlength{\tabcolsep}{4pt}
\begin{tabular}{lrrrr}
\toprule
Domain            & Refused & False cert. & Loop-fail & $n$ \\
\midrule
Drone             & 0 & 0 & 5 & 5 \\
Biochem           & 0 & 0 & 5 & 5 \\
Astrophysics      & 0 & 0 & 5 & 5 \\
Electrical        & 0 & 0 & 5 & 5 \\
Drug formulation  & 0 & 0 & 5 & 5 \\
\midrule
\textbf{Overall}  & \textbf{0} & \textbf{0/25} & \textbf{25} & \textbf{25} \\
\bottomrule
\end{tabular}
\end{table}

\begin{figure*}[t]
\centering
\resizebox{0.98\linewidth}{!}{%
\begin{tikzpicture}[x=1cm,y=1cm]

  \draw[thick, color=BlueGray] (0,0) -- (0,5.2);
  \draw[thick, color=BlueGray] (0,0) -- (5.0,0);
  \foreach \y/\lbl in {0/0,1/20,2/40,3/60,4/80,5/100}{
    \draw[BlueGray!30, thin] (0,\y) -- (5.0,\y);
    \node[left,font=\scriptsize\color{BlueGray}] at (-0.1,\y) {\lbl\%};
  }

  \node[rotate=90,font=\small\bfseries\color{BlueGray}]
      at (-1.45,2.5) {Feasible certification rate};

  \draw[BlueGray,thick] (0.55,0) -- (1.10,0);
  \node[above,font=\scriptsize\bfseries\color{BlueGray}] at (0.825,0.10) {n/a};

  \fill[BlueGray!55] (2.05,0) rectangle (2.60,3.30);
  \node[above,font=\scriptsize\bfseries\color{BlueGray}] at (2.325,3.30) {66\%};

  \fill[BlueGray!85] (3.55,0) rectangle (4.10,4.40);
  \node[above,font=\scriptsize\bfseries\color{BlueGray}] at (3.825,4.40) {88\%};

  \node[below,font=\scriptsize\color{BlueGray},text width=1.4cm,align=center]
      at (0.825,-0.15) {Llama\\3.1 8B};
  \node[below,font=\scriptsize\color{BlueGray},text width=1.4cm,align=center]
      at (2.325,-0.15) {Llama\\3.3 70B};
  \node[below,font=\scriptsize\color{BlueGray},text width=1.4cm,align=center]
      at (3.825,-0.15) {Gemini\\2.5 Flash};

  \node[font=\small\bfseries\color{BlueGray}]
      at (2.5,5.65) {(a) Feasible certification rate};

  \begin{scope}[xshift=6.8cm]

    \draw[thick,color=BlueGray] (0,0) -- (0,5.2);
    \draw[thick,color=BlueGray] (0,0) -- (5.0,0);

    \foreach \y/\lbl in {0/0,1/20,2/40,3/60,4/80,5/100}{
      \draw[BlueGray!30,thin] (0,\y) -- (5.0,\y);
      \node[left,font=\scriptsize\color{BlueGray}] at (-0.1,\y) {\lbl\%};
    }

    \node[rotate=90,font=\small\bfseries\color{BlueGray}]
        at (-1.45,2.5) {False certification rate (bare validator)};

    \fill[BlueGray!30] (0.55,0) rectangle (1.10,0.04);
    \draw[BlueGray,very thick] (0.55,0.04) -- (1.10,0.04);
    \node[above,font=\scriptsize\bfseries\color{BlueGray}]
        at (0.825,0.15) {0\%$^\dagger$};

    \fill[RedDark!55] (2.05,0) rectangle (2.60,0.20);
    \node[above,font=\scriptsize\bfseries\color{RedDark}]
        at (2.325,0.20) {4\%};

    \fill[RedDark!80] (3.55,0) rectangle (4.10,0.40);
    \node[above,font=\scriptsize\bfseries\color{RedDark}]
        at (3.825,0.40) {8\%};

    \node[below,font=\scriptsize\color{BlueGray},text width=1.4cm,align=center]
        at (0.825,-0.15) {Llama\\3.1 8B};

    \node[below,font=\scriptsize\color{BlueGray},text width=1.4cm,align=center]
        at (2.325,-0.15) {Llama\\3.3 70B};

    \node[below,font=\scriptsize\color{BlueGray},text width=1.4cm,align=center]
        at (3.825,-0.15) {Gemini\\2.5 Flash};

    \node[font=\small\bfseries\color{BlueGray}]
        at (2.5,5.65) {(b) Safety (bare validator)};

  \end{scope}

\end{tikzpicture}%
}

\caption{\textbf{Capability governs feasible certification rate but not safety.}
(a) Feasible certification rate rises with proposer capability: Llama~3.1~8B
never closes the loop (n/a; not run on feasible goals), Llama~3.3~70B reaches
66\%, and Gemini~2.5~Flash reaches 88\%.
(b) False certification rate under the bare validator: 4\% for Llama~3.3~70B
and 8\% for Gemini. Llama~3.1~8B ($^\dagger$) shows 0\% not because the contract
caught it but because it never issues a verdict at all: every trial ends in
loop-closure failure. For Gemini and Llama~3.3~70B, the structural contract
yields 0\% false certifications. Capability scales feasible certification
rate; the structural contract, not the proposer, determines whether a verdict
once issued can be wrong.}

\label{fig:ladder}
\end{figure*}
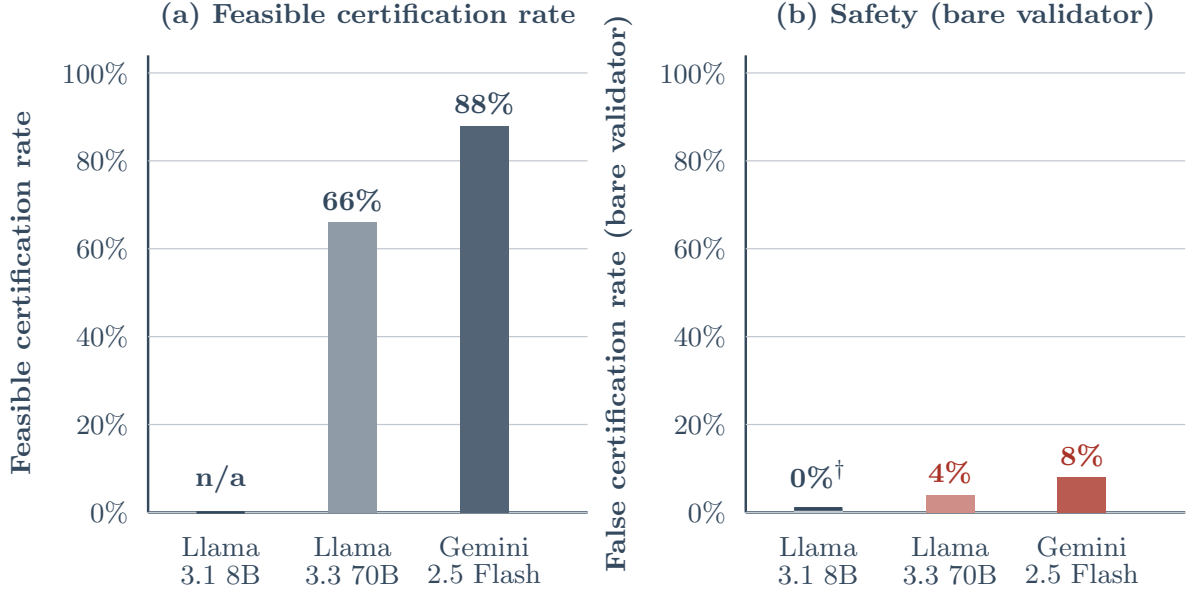

Taken together with the cross-model result, the capability ladder reads cleanly from top to bottom (Fig.~\ref{fig:ladder}): Gemini closes the loop reliably, Llama~3.3~70B stalls on the harder multi-tool domains, and Llama~3.1~8B never closes it at all, yet neither of the three ever forges a certificate. Capability governs whether the system produces a verdict; it has no bearing on whether a verdict, once produced, can be wrong, because the authority to certify never passed through the proposer. A loop-closure supervisor (a model-agnostic component that detects a stalled turn after an unsatisfied rejection and re-prompts the tool call) would plausibly recover much of the lost feasible certification rate for weaker proposers without touching the engine's authority. We have not implemented or evaluated it; it is a clearly bounded direction for future work.

\subsection*{What the verification and the language model each contribute}

Two fairness baselines isolate where the value of the system actually lies, and both make the contribution of PHACT's components more precise rather than larger.

The first asks whether PHACT's safety advantage is merely that PHACT is allowed to decline. We therefore re-ran the bare model with a prompt that explicitly permits it to answer ``impossible'' when no valid design exists (no engine, otherwise identical). Permitting refusal does reduce false certifications: the bare-but-permitted model falsely certifies 2 of 25 impossible goals rather than 7 of 25 when forbidden. But it does not eliminate them. Under the structural contract, false certifications drop to zero for every model tested. The prompt cannot supply what the contract provides: the permitted bare model has no way to verify its own refusals, still falsely certifies goals it cannot catch, and over-refuses four feasible goals it should have designed. None of its answers are checked. PHACT's value is therefore not that it can decline, which a prompt can approximate, but that every certificate it issues is verified and, under the structural contract, cannot be forged.

The second baseline asks what the language model contributes that a solver does not. For the domains whose design is a direct inversion of a closed-form law, the resistor--capacitor cutoff $R = 1/(2\pi f_c C)$ and the chirp mass $M_c = (m_1 m_2)^{3/5}/(m_1+m_2)^{1/5}$, a thin deterministic solver with no language model certifies every target we tried (7 of 7 RC filters, 7 of 7 chirp masses). Where a closed form exists, the physics needs no model, and a parser plus the engine would suffice. We state this plainly because it locates the language model's actual marginal contribution: not the physics, which a solver computes exactly, but the natural-language front end and the generality of operating across five unrelated domains, from DNA thermodynamics to colloidal stability, without per-domain solver code. The value of PHACT is best understood as a method for safely using an unreliable proposer in the open-ended setting where no solver exists, with the closed-form domains serving as a control that confirms the engine, not the model, is the source of physical correctness.

\subsection*{When the form of the feedback matters}

The engine returns, on failure, a violation report and a targeted correction. Does a richer, structured correction help the model converge faster? We answer with a controlled ablation holding the model, goals, and physics fixed, varying only the feedback form (weak: violation magnitude only; strong: closed-form correction; structural: the root-input correction with the dependency path), across three domains (Table~\ref{tab:feedback}). The answer is conditional. On the RC filter, whose corrective inverse is closed-form ($R = 1/(2\pi f_c C)$), all three forms converge in one to one-and-a-half iterations, the model computes the fix itself and the feedback form is immaterial. On drug formulation, whose feasible region is large, the first proposal usually certifies and again the form barely matters (1.20 vs 1.00 iterations). Only on DNA dual-constraint design, where two coupled constraints share parents and no closed-form inverse exists, does structural feedback meaningfully reduce iterations, from 2.50 (weak) to 1.70, a roughly 30\% improvement, with strong feedback intermediate at 2.22. Structural feedback earns its complexity precisely when the corrective inverse requires joint optimization the model cannot perform unaided; elsewhere, the choice is immaterial.

\begin{table}[t]
\centering
\caption{\textbf{Feedback-form ablation (Gemini 2.5 Flash; mean iterations to certify).} Weak: violation magnitude only. Strong: closed-form correction. Structural: root-input correction with dependency path. Feedback form matters only for the coupled domain (DNA), where no closed-form inverse exists; for a closed-form inverse (RC) or a large feasible region (drug) it is immaterial.}
\label{tab:feedback}
\setlength{\tabcolsep}{4pt}
\begin{tabular}{lrrr}
\toprule
Domain (correction structure)       & Weak & Strong & Struct. \\
\midrule
RC filter (closed-form inv.)        & 1.00 & 1.50 & 1.10 \\
Drug form.\ (large region)          & 1.20 & 1.20 & 1.00 \\
DNA dual-constraint (coupled)       & 2.50 & 2.22 & 1.70 \\
\bottomrule
\end{tabular}
\end{table}

\section*{Related work}

\paragraph{Tool-augmented and program-aided models} Language models can call external tools or offload computation to a code interpreter\cite{schick2023toolformer,yao2023react,gao2023pal,chen2022program}. These verify the \emph{execution of code}, not the \emph{physical admissibility} of a design; a program that runs cleanly can still describe a device that violates physical law. PHACT's validators check the latter, and our central finding, that the validator must derive rather than accept the certified quantity, is a constraint on \emph{how} such a tool must be wired, not merely \emph{that} one is present.

\paragraph{Self-correction and verification} A model can criticize and revise its own output\cite{madaan2023selfrefine,shinn2023reflexion}, and external verifiers can score candidates\cite{cobbe2021verifiers}. Self-verification inherits the blind spots that produced the error\cite{huang2024selfcorrect}; PHACT's verifier is a deterministic engine with which the model cannot argue. The forgery we identify is precisely a case where a naive external verifier is still defeated (not by being wrong, but by being asked the wrong question) which is why the input/output contract matters.

\paragraph{Neurosymbolic systems and structured verification} Combining neural generation with symbolic reasoning is a long-standing approach\cite{garcez2019neurosymbolic,marcus2020next}. We take a known structural model of a physical mechanism, express it as a directed dependency graph whose input/output structure makes certification non-forgeable, and use that graph to generate targeted corrections when a design fails. The contribution is in how the graph is wired relative to the proposer, not in inferring the graph from data; we take the physical laws as given and use their dependency structure as an interface contract.

\paragraph{Hallucination, abstention, and physics-informed learning} Models produce fluent false statements\cite{ji2023survey,maynez2020faithfulness}, motivating calibrated abstention\cite{kadavath2022know,ren2023outofdistribution}; PHACT's three-valued output is a hard form of abstention grounded in an external check rather than self-estimated confidence. Physics-informed neural networks fold the governing equations into the training loss as a soft penalty\cite{raissi2019pinn,karniadakis2021pinnreview,sanchez2020learning}; there the law is internalized and can still be violated at inference, and the network always emits a value. PHACT keeps the law outside the model as a hard check at inference: the model's only authority is to propose, never to certify. The distinction is not which approach knows more physics but where the law sits relative to the output: inside the model as a training signal, or outside it as a runtime gate. The two are complementary rather than competing. A physics-informed surrogate trained on a domain whose governing law has no closed form could slot directly into PHACT as the engine\cite{raissi2019pinn}, extending certification to regimes where ordinary code cannot compute the forward model exactly. In that configuration it is still the certification contract, not the soft constraint, that removes the confident assertion of the impossible; the surrogate supplies the forward model, and the contract determines who may assert the result.

\section*{Discussion}

The improvement we find most significant is not its size but its source, and what it costs to make it sound. The bare model and PHACT are the same model; we trained nothing and collected no data. We added a loop in which the model may not assert a quantity until a deterministic engine has confirmed it. That single change moves the system from one that is wrong roughly half the time, and confidently designs the impossible, to one that certifies the large majority of valid designs and refuses the rest.

But a certificate is only as good as the contract that issues it. The central finding is how that contract can be forged and what closes the hole. When the validator asks the proposer to supply the quantity under test, the proposer can satisfy it by changing a quantity the prompt declared fixed; the engine then certifies correct physics for the wrong design. We showed this occurs readily in the domains most susceptible to it, and that it is worst under the greedy decoding with which models are usually run. The fix is structural and, once seen, simple: derive the certified quantity from the fixed inputs so the proposer cannot write it. Under this contract, the forgery is impossible by construction, and the empirical results bear it out across models, temperatures, and even a deliberately faulted engine, zero forged certificates in eighty adversarial trials. The guarantee is structural rather than statistical, which is why it does not depend on how the model is decoded: the authority to assert never passes through the sampler. This sharpens a familiar intuition. Reliability is not only a property of how good the predictor is; it is a property of what the predictor is allowed to assert without a check it cannot author.

The five domains and three proposers evaluated here are not the limits of the architecture; they are a first instantiation of it with closed-form engines. Each axis admits direct extension. On the proposer side, the experiments show that loop-closure is a meta-cognitive capacity separable from physical knowledge: Llama~3.3~70B correctly parses the engine's correction and sometimes restates the right value, yet fails to resubmit it as a tool call, ending with no verdict. The safety property is untouched by this failure mode, the model never forges, and it points to a concrete next step that does not require a better model: a model-agnostic loop-closure supervisor that detects a stalled natural-language turn after an unsatisfied rejection and re-prompts the tool call. With capable proposers, the loop already closes reliably; the supervisor is the path to making that hold for smaller, fully open models and to broadening access further. On the engine side, the five domains here all have closed-form governing equations, which is why certification is fast and exact. The architecture places no constraint on this: any deterministic forward model, including a physics-informed surrogate for a domain whose law has no closed form, can serve as the engine. The \emph{unknown} verdict is the system's honest response when the engine's competence runs out, and widening that competence is a separate axis from the certification contract itself, one that the community can pursue domain by domain without changing the safety structure. The one boundary that is not an evaluation limit but a property of any PHACT certificate is that soundness is relative to the engine's model of the physics: the certificate says this design satisfies these equations, not that it satisfies physical reality in full. We regard this as the correct scope for any computable certificate, and stating it explicitly is what makes the guarantee trustworthy.

The certification path contains no machine learning and no proprietary model: the engines are ordinary code, and the released archive regenerates every result on commodity hardware without credentials, a network, or a language model. A student or a working scientist can put a candidate design, a DNA sequence intended to be stable at a target temperature, a filter sized for a cutoff, a nanoparticle formulation meant to resist aggregation, through the same governing physics we use here, and learn whether it is admissible, infeasible, or outside the engine's competence, before committing bench time or an HPC allocation to it. The system does not replace high-fidelity simulation; it screens against the first-principles law cheaply, so that expensive resources are spent only on designs that have already cleared it.

The separation of authority instantiated here, a learned proposer paired with a deterministic arbiter the proposer cannot override, may generalize beyond static physical design. We make no claim about real-time control or perception, which this evaluation does not touch; but the same input/output contract that prevents forgery on a fixed-mass chirp-mass goal is the same contract one would want wherever a model's output feeds an external check it cannot author.

Reliable physical reasoning is not a property of a better predictor; it is a property of a predictor placed inside a loop whose verdict it cannot author. PHACT is one way to build that loop.

\section*{Methods}

\paragraph{Models and decoding} The primary proposer, in both the bare and PHACT conditions, is Gemini 2.5 Flash via Vertex AI, decoded at temperature 0; the same model is used for both conditions so that the engine is the only variable. The cross-model experiment uses Llama~3.3~70B via Vertex AI Model-as-a-Service, also at temperature 0; its bare baseline uses the same API. A capability-floor proposer, Llama~3.1~8B (an order of magnitude smaller), was run on the impossible-goal suite served locally via Ollama, entirely on-device with no external API. The high-temperature ablation repeats the adversarial impossible suite at temperature 1. No model was fine-tuned. Determinism at temperature 0 was confirmed by byte-identical tool-call trajectories on repeated runs. The proposer is orchestrated with the Google Agent Development Kit, which lets the model decide when to invoke tools; non-Gemini models route through the ADK's LiteLLM backend. A run ending in error is classified at source as an infrastructure fault (rate limit, authentication, server 5xx, or stream timeout), which is retried and never scored, or a loop-closure failure (no terminal verdict), which is a property of the proposer and is reported as such. We evaluate three models in total; broader generality would be firmer with additional capable proposers run end to end.

\paragraph{The engine and the structural contract} Each domain engine is a deterministic implementation of its governing equations, containing no machine learning, calibrated against published values (DNA melting temperature within $\pm1\,^\circ$C of SantaLucia 1998; GW150914 chirp mass within $\pm1\,M_\odot$ of the LIGO/Virgo catalog) to confirm correct implementation, a method check, not a performance result. We do not score the system on famous published quantities, because a model trained on the reporting papers recalls rather than computes them; all performance claims rest on design and impossible goals where memorization cannot help. For each domain we declare a structural contract: a directed dependency graph with two kinds of input nodes, goal-bound quantities and proposer-controlled design variables, followed by derived quantities and an outcome node for the certified quantity; the edges are the same closed-form equations as the validator, verified to reproduce the bare-tool values exactly. For fixed-system verification goals, the bound quantities are read from the goal state and are not accepted as model arguments. The certification contract is selected by tool availability: a goal that fixes a system and asks to verify a derived property is certified only through the structural (derive-from-inputs) check, with the answer-submitting validators removed and the fixed quantities bound outside the model's control, so the proposer cannot change the system under test or supply the value under test. The forgeability ablation toggles this single variable, answer-submitting validators (``bare'') versus the structural check, on identical goals.

\paragraph{Goals and scoring} For each domain we authored ten feasible and five impossible goals; an impossible goal fixes an over-constrained input set that admits no valid design. The goals, engines, and feasible/impossible boundary are author-defined; there is no held-out external goal set, so the evaluation is a closed-world test of the architecture rather than a benchmark against an independent standard, and the rates should be read in that light. A bare answer is scored by parsing its parameters and evaluating them with the same engine. A PHACT run is scored by its returned outcome. For feasible goals, success is certification, and feasible certification rates are reported over all 50 authored goals, with timeouts shown explicitly rather than excluded. For impossible goals, success is the verdict impossible. Infrastructure faults are retried and not scored; wrong verdicts and loop-fails are reported separately and never counted as successes; a false certification on an impossible goal is counted as a failure to refuse, never excluded.

\paragraph{Statistics} Wilson score 95\% confidence intervals are reported for the genuinely empirical-rate rows, where a rate is being estimated from finite samples. We deliberately report the central forgeability contrast descriptively rather than through a significance test. The structural contract's zero forgery result is a property of the interface, the certified quantity is computed and cannot be authored by the model, not a difference inferred from sampling; a $p$-value would mischaracterize a by-construction guarantee as a statistical one. The structural zero rows are therefore presented as implementation checks under the archived adversarial conditions: no forged certificate was observed. For the bare contract, which is a genuine empirical rate, we report the per-condition proportions and their intervals. Given the small per-cell sizes ($n=15$ to $50$), we read the intervals, not point estimates, as the result, and we flag where two intervals overlap (so a difference is directional rather than significant) rather than asserting a difference the data do not support.

\paragraph{Reproducibility} The complete framework is released as a public archive: the physics engines, the structural contracts (the input/output dependency graphs), the full goal corpus, the experiment driver, the engine-calibration scripts, and a table-regeneration pipeline. The archive is the soundness claim made checkable: every table in this paper regenerates from the archived model outputs and the deterministic engines \emph{without credentials, network, or a language model}, so a reader can reproduce the full analysis pipeline and verify the physics exactly, end to end, with no API key. What the archive does not reproduce is the elicitation of model behavior, which depends on the hosted models and provider-side non-determinism; the certificate-checking pipeline, which is where soundness lives, is fully deterministic and fully reproduced.

\section*{Acknowledgements}

NV and IDS thank the Karlsruhe Institute of Technology (KIT) and the University of Stuttgart for institutional support. IDS further acknowledges support from KIT's Excellence Strategy via the Young Investigator Group Preparation Program. Heysuvi Labs, LLC operated in part under a technology transfer initiative of the University of Stuttgart during the development of this work. This work also benefited from consulting and in-kind support provided through the NVIDIA Inception and Google for Startups programs.

\section*{Funding}

This work was funded entirely by Heysuvi Labs, LLC. No external grant funding was received.

\section*{Competing interests}

NV and IDS are employees of Heysuvi Labs, LLC, which funded this work. The authors declare no other competing interests.

\section*{Ethics approval and consent to participate}

Not applicable.

\section*{Consent for publication}

Not applicable.

\section*{Data and code availability}

The goal corpus, engine outputs, and all data required to reproduce the tables and figures in this paper are available at \url{https://github.com/nakulvyas21/phact-reproducibility}. The physics engines, structural contracts, experiment driver, and table-regeneration pipeline are publicly available at \url{https://github.com/nakulvyas21/phact-reproducibility}. A production user interface based on the PHACT workflow is available at \url{https://heysuvi.com/ae210}. Every result in this paper regenerates from the archived outputs and the deterministic engines without credentials, a network connection, or a language model.

\section*{Author contributions}

NV conceived and designed the study, developed the framework and physics engines, ran the experiments, and wrote the manuscript. IDS provided scientific guidance, reviewed and approved the final manuscript.

\bibliography{references}

\end{document}